# Advancing Wood Identification in the Philippines: Utilizing the Xylorix Platform for Efficient AI Model Development and Deployment for Five Key Species

Rosalie C. Mendoza* (College of Forestry and Natural Resources, University of the Philippines Los Banos), Vivian C. Daracan (College of Forestry and Natural Resources, University of the Philippines Los Banos), Arlene D. Romano (College of Forestry and Natural Resources, University of the Philippines Los Banos), Ronniel D. Manalo (College of Forestry and Natural Resources, University of the Philippines Los Banos), Xin Jie Tang (Agritix), Yi Hong Wong (Agritix), Yong Haur Tay (Agritix)

## Abstract

Illegal logging and timber trade continue to pose significant challenges in the Philippines, where accurate wood species identification is essential for enforcement but limited by the need for specialised equipment and expertise. This study aims to evaluate whether AI models for macroscopic wood identification can be developed and deployed by wood scientists without programming expertise using the Xylorix platform, focusing on five Philippine hardwood species: Mangium (*Acacia mangium* Willd.), Rain Tree [*Samanea saman* (Jacq.) Merr.], Banuyo (*Wallaceodendron celebicum* Koord.), Tindalo [*Afzelia rhomboidea* (Blanco) Vidal], and Ipil [*Intsia bijuga* (Colebr.) O. Kuntze]. Binary classifiers were trained on 10,663 verified cross-section images from 260 specimens and evaluated using specimen-level mean scoring to mirror operational field conditions. Area Under the ROC Curve (AUC) values ranged from 0.969 (Ipil) to 1.000 (Mangium), and Average Precision (AP) values ranged from 0.589 (Samanea) to 1.000 (Mangium). Four of five species achieved AA grade (AUC and AP both \geq 0.90); Rain Tree received AE (AUC \geq 0.90, AP < 0.60) due to AP compression from its small positive test set (3 specimens). All five classifiers rank their target specimens above non-target specimens with near-perfect fidelity. Specimen-level error analysis revealed 9 false negatives from Ipil, primarily stemming from localized image artifacts and 3 false positives for Rain Tree and 1 false positive for Tindalo caused by shared tribal-level anatomical traits. These findings demonstrate that Xylorix non-programmers can leverage the Xylorix platform to construct operationally reliable wood identification models suitable for field deployment at supply chain checkpoints.

## 1. Introduction

Accurate wood identification is essential in the Philippines for conservation efforts and combating illegal logging, given the archipelago's rich biodiversity and the persistent challenges posed by the illicit timber trade. Traditional wood identification methods, which rely on macroscopic and microscopic anatomical analysis, are time-consuming, require specialized expertise, and hinder rapid enforcement and large-scale monitoring. Consequently, artificial intelligence programs offer a promising avenue to enhance the efficiency and accessibility of wood identification, thereby strengthening regulatory compliance and conservation initiatives (Urbano et al., 2023). Such AI-driven tools are proving invaluable by expediting identification and expanding their applicability to field settings and complex anatomical structures (Nieradzik et al., 2023; Ravindran et al., 2021). The integration of computer vision and machine learning techniques, particularly through platforms such as Xylorix, has demonstrated significant potential to deliver accurate, rapid, and scalable identification of timber species, thereby addressing the limitations of traditional methods (B. et al., 2021; Hwang & Sugiyama, 2021; Ravindran et al., 2021). These advanced systems, often employing convolutional neural networks, leverage extensive databases of wood characteristics to classify species with high reliability, thereby enabling more effective enforcement against illegal logging and supporting international agreements such as CITES (Ravindran et al., 2021; Urbano et al., 2023). This technological shift is critical for nations like the Philippines, where unique local variations in wood species necessitate tailored identification tools, despite global efforts to standardize these systems (Ravindran et al., 2021; Urbano et al., 2023).

To address these challenges, the Xylorix platform brings the expertise of a wood anatomist to the palm of the end user such as forestry officer, customs inspector, timber trader, and wood product manufacturer. Using advanced AI and machine learning to analyze wood cross-sections, Xylorix delivers fast, reliable species identification with only a few basic tools: an illuminated macro-lens, a utility knife, a thumb protector, and a smartphone. This minimal setup is deliberately designed for real-world field conditions, where operators make quick cuts in varied lighting rather than preparing perfectly polished laboratory specimens. As a result, the quality of individual images can vary, and the platform handles this by capturing multiple images per specimen and averaging the model's predictions. This specimen-level averaging, described in Section 3, provides a stable identification that reflects the specimen's true identity rather than the quality of any single image, mirroring how end users would rely on consistent results across multiple field captures. Xylorix thus eliminates the need for bulky manuals, lab equipment, or on-site expert consultation, making accurate wood identification accessible in forests, ports, sawmills, and beyond.

This study evaluates whether binary deep learning classifiers for five Philippine hardwood species, Mangium (*Acacia mangium*), Rain Tree (*Samanea saman*), Banuyo (*Wallaceodendron celebicum*), Tindalo (*Afzelia rhomboidea*), and Ipil (*Intsia bijuga*), can achieve operational reliability when developed by wood scientists without programming expertise using the Xylorix platform. The primary research objectives involve quantifying per-species classification performance via specimen-level mean scoring, characterizing cross-species confusion patterns relative to their anatomical bases at 24× magnification, and designing species-specific decision thresholds to counteract sensitivity-specificity imbalances during field deployment.

# 2. Materials and Methods

## 2.1 Xylorix Platform Overview

This section examines the architectural design and operational principles of the Xylorix platform, elucidating how its integrated AI models enable rapid and robust identification of wood species. Like the other AI-aided models, it leverages a sophisticated combination of image processing, machine learning algorithms, and a comprehensive database of wood anatomical features to deliver precise classifications (Liu et al., 2024).

As shown in Figure 1, the Xylorix platform is designed to simplify the integration of artificial intelligence into wood science workflows. It effectively hides the complexity of AI model development, allowing wood scientists to focus on specimen analysis and domain expertise to develop effective AI models rather than on the computational processes and software technologies required to build them. Through a coordinated system of mobile applications, web-based tools, and cloud infrastructure managed by the Xylorix team, Xylorix Platform enables seamless collaboration between wood scientists and technical teams.

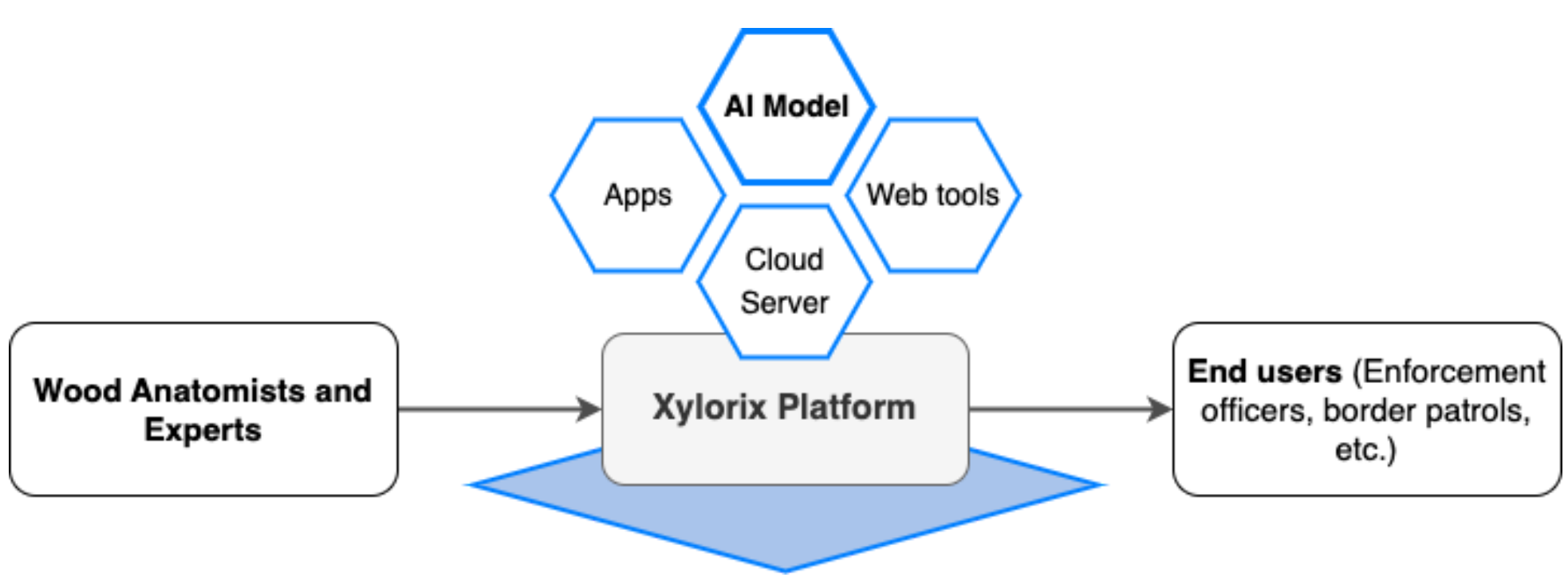

*Figure 1. Overview of Xylorix Platform*

As part of the coordinated system, Xylorix provides an integrated suite of applications that streamlines the entire model development process. These include tools for data collection (Xylorix Harvester), image verification (Xylorix Toolkit Verifier), model evaluation (Xylorix Toolkit Insider and Xylorix Insider), and deployment (Xylorix Inspector).

As illustrated in Figure 2, the Xylorix model-building workflow comprises three main phases: data collection and verification, model building and evaluation, and model deployment. Automation within these processes reduces human error, ensures data consistency, and accelerates the development of reliable AI models. The operational details of each phase are elucidated in the subsequent paragraphs.

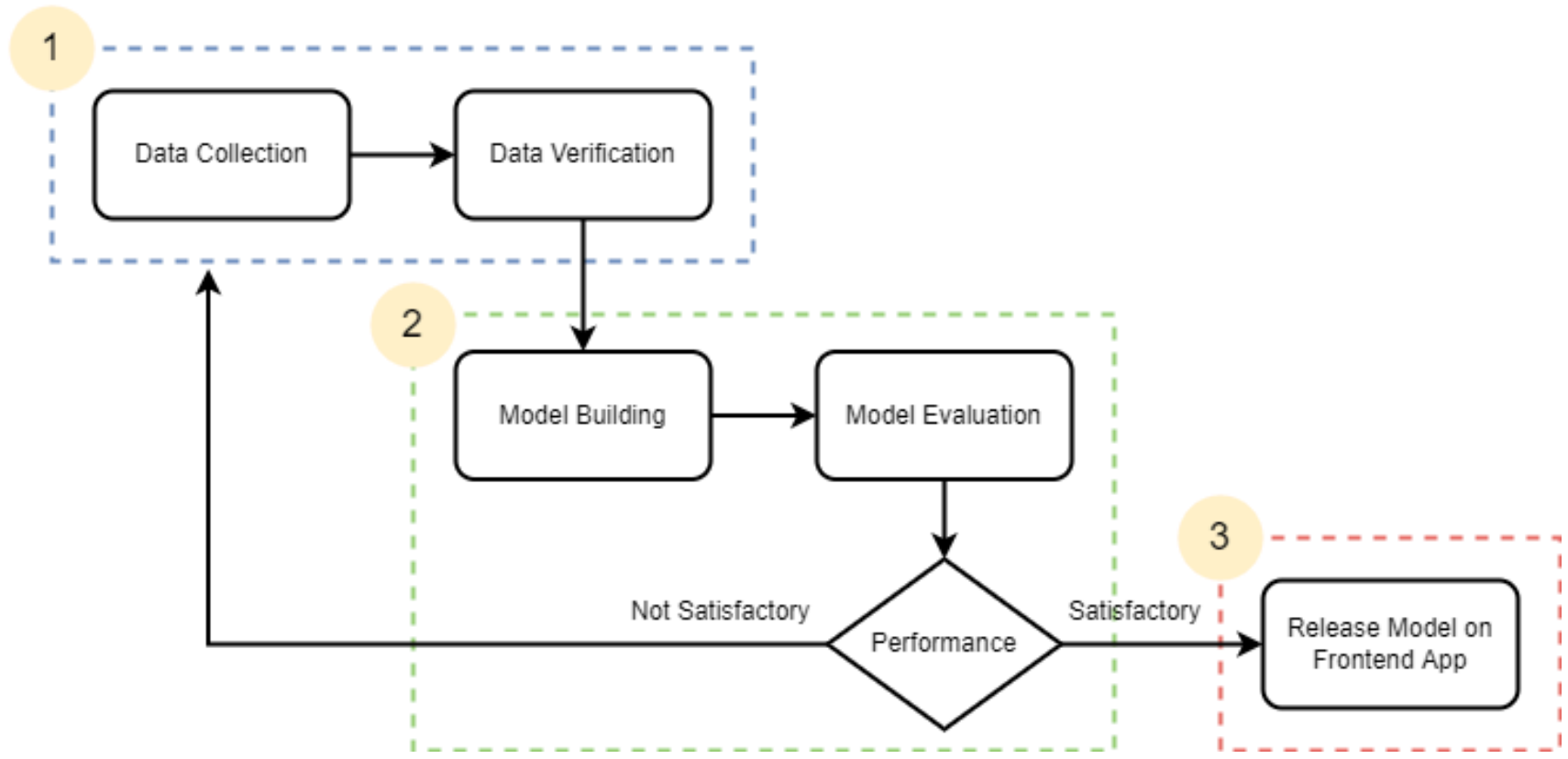

*Figure 2. Flow chart of Xylorix Model Building Process*

In the initial phase, the UPLB team members acquire macroscopic wood specimen images using a dedicated data collection application, Xylorix Harvester. This application automatically organizes the captured images by group, species, and specimen, subsequently storing them on the cloud server, thereby reducing the risk of human error throughout the process. The uploaded images then undergo a verification stage, in which each image is meticulously reviewed by wood anatomists using a specialized web-based verification tool, Xylorix Verifier. This verification process serves as a crucial gatekeeper, ensuring that only images that meet the stringent criteria for developing robust AI models are used.

In the second phase, the Xylorix team proceeds to develop the AI models using the verified data. Upon creating the initial model version, the team introduces a different set of testing instruments, namely the Xylorix Toolkit and Xylorix Insider. These tools enable the UPLB team to systematically evaluate the AI models' performance and analyze their efficacy in executing identification tasks. Based on the model performance results, the UPLB team and the Xylorix team jointly decide whether the models are performing satisfactorily. Should the models be deemed underperforming, the team will return to the initial phase to acquire additional data or adjust the model training parameters to enhance or refine the AI models until satisfactory performance is achieved.

In the third phase, once the AI models are rated ready to use as ascertained by the UPLB team, the Xylorix team will release the models into the front-end apps, such as the Xylorix Inspector, and make them accessible to the end users.

As part of the key innovation of the Xylorix Platform, the Xylorix team also developed the WIDK-24X01, a specially designed 24x magnification macro lens kit with built-in rechargeable illumination that can be mounted on smartphones, as shown in Figure 3. This tool allows users to capture high-quality macroscopic images of wood cross-sections under consistent lighting and magnification conditions for identification and verification in the field. The portability and ease of use of this setup make it highly suitable for field applications, including forests, ports, and sawmills. The UPLB team members are also equipped with the same setup (mobile phones with macro lenses) during the model-building process to ensure consistency in magnification and image quality for model building and, eventually, for field application by end users as well.

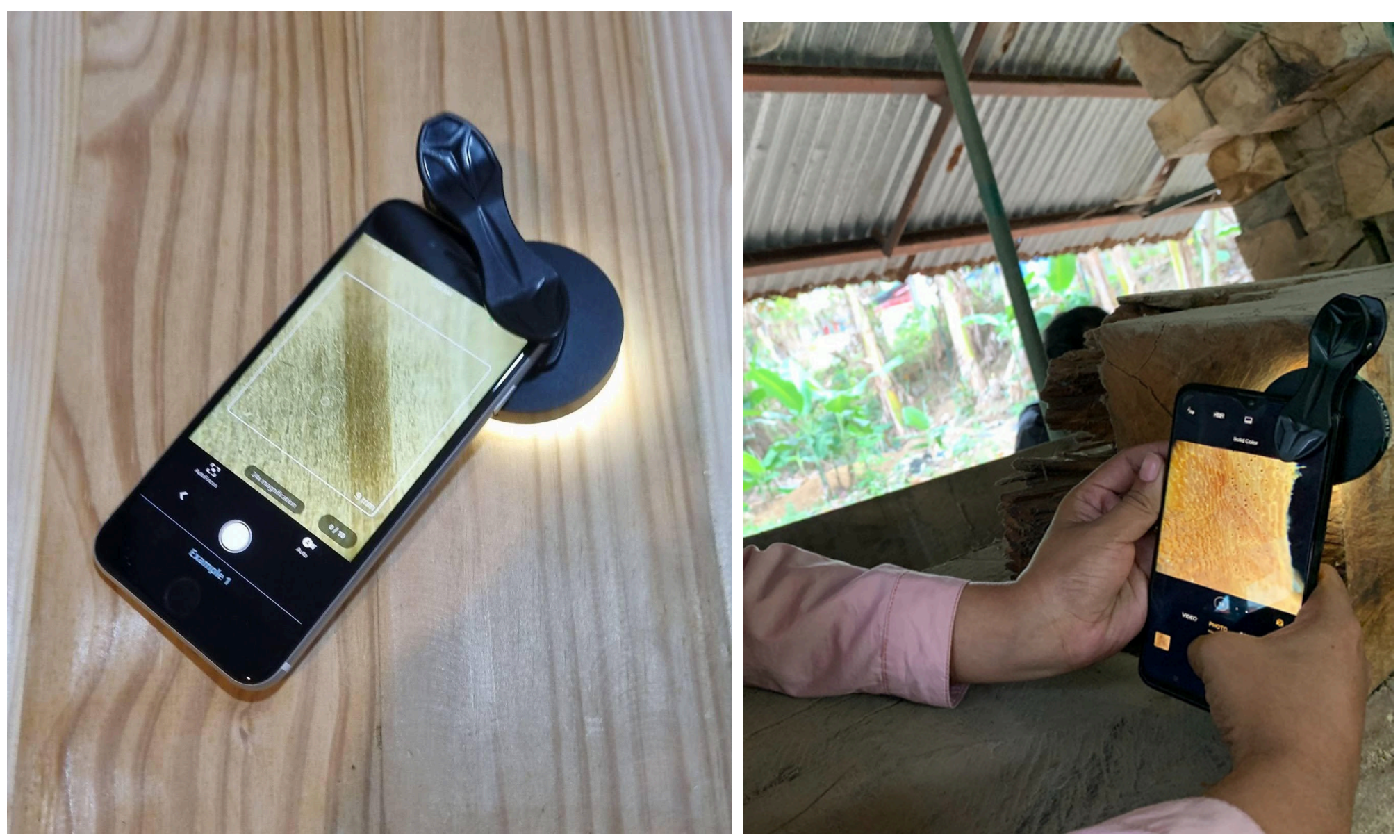

*Figure 3. Smartphone mounted with the WIDK-24X01 macro lens.*

As shown in Table 1, the five key species considered in this study were Mangium (*Acacia mangium* Willd.), Rain Tree [*Samanea saman* (Jacq.) Merr.], Banuyo (*Wallaceodendron celebicum* Koord.), Tindalo [*Afzelia rhomboidea* (Blanco) Vidal], and Ipil [*Intsia bijuga* (Colebr.) O. Kuntze]. These species were selected based on the following criteria: (1) species usually found in impounding sites of the Department of Environment and Natural Resources or DENR or used commercially, (2) species with prominent similarities, and (3) species with a considerable number of wood specimens available.

All five species belong to the Fabaceae family. Tindalo and Ipil share the Detarioideae subfamily and Afzelieae tribe, while Mangium, Rain Tree, and Banuyo belong to Caesalpinioideae but different tribes. This taxonomic structure provides a natural hierarchy for evaluating whether the AI models can learn subfamily-level and tribe-level anatomical distinctions from macroscopic images.

Ipil and Tindalo are both premium hardwood species that share a high anatomical uniformity, making them difficult to distinguish at low magnifications. As described in Meniado et al. (1975), both species exhibit vessels typically arranged in solitary and in radial multiples of 2-3, accompanied by vasicentric, aliform, and aliform confluent parenchyma. Ipil generally features moderately large pores containing distinctive yellowish sulfur-like deposits, whereas Tindalo exhibits moderately small pores commonly filled with reddish deposits. While both species have boundary parenchyma, the wood of Tindalo is known for having finer texture and a darker wine-red color compared to the orange-brown color of Ipil. Physically, both species are characterized as hard and heavy.

Similarly, Rain tree is frequently compared to Banuyo due to their overlapping characteristics. Rain tree is uniquely identified by its dull luster and very thick vasicentric parenchyma, making its moderately large pores readily visible to the naked eye. In contrast, Banuyo is distinguished by an echelon arrangement of pores and narrower vasicentric parenchyma compared to Rain tree. Mangium is differentiated by its smaller, moderately numerous pores that are often arranged in radial multiples of 2-4 or in tangential multiples of 2, accompanied by thinner to indistinct vasicentric parenchyma. Physically, Banuyo exhibits a light golden brown color while Rain tree and Mangium are distinguished by their lighter whitish to light brown heartwood (Meniado et. al, 1975).

*Table 1. The five key Philippines species.*

| Common Name | Scientific Name | Family | Subfamily | Tribe |
|---|---|---|---|---|
| Mangium | *Acacia mangium* Willd. | Fabaceae | Caesalpinioideae | Acacieae |
| Rain tree | *Samanea saman* (Jacq.) Merr. | Fabaceae | Caesalpinioideae | Ingeae |
| Banuyo | *Wallaceodendron celebicum* Koord. | Fabaceae | Caesalpinioideae | Ingeae |
| Tindalo | *Afzelia rhomboidea* (Blanco) Vidal | Fabaceae | Detarioideae | Afzelieae |
| Ipil | *Intsia bijuga* (Colebr.) O. Kuntze | Fabaceae | Detarioideae | Afzelieae |

The IAWA (International Association of Wood Anatomists) List of Microscopic Features for Hardwood Identification (IAWA Committee, 1989) was also used to compile the wood anatomical feature sets for each species from the InsideWood database (Wheeler, E.A. 2011). Appendix A provides the full IAWA hardwood feature codes for each species, and Table 2 identifies which features are observable at the 24× magnification provided by the WIDK-24X01 macro-lens used in this study.

All five species share a core set of Fabaceae anatomical features: diffuse-porous arrangement (5), vessels in radial multiples (10), simple perforation plates (13), alternate intervessel pits (22), vestured pits (29), vessel-ray pitting (30), axial vasicentric and aliform parenchyma (79, 80), axial parenchyma cell type/strand length (91, 92)  rays ≤4 cells wide (97), presence of procumbent ray cells (104), number of rays per mm (97), and presence of prismatic crystals (136). Other than these common features of all the five species, many of the features are also characteristics of more than two species included in this study. These shared traits create a baseline of anatomical uniformity across the species.

*Table 2. Discriminating features observable at the macroscopic level (24× magnification).*

| IAWA Feature | Observability at 24x | Species Variation | Diagnostic Utility |
|---|---|---|---|
| Growth ring distinctness (1/2) | Partial | Variable in all species; limited diagnostic value | - |
| Wood diffuse porous (5) | Yes | None | - |
| Vessels in radial multiple (10) | Yes | Frequency of occurrence variable in all species | - |
| Mean tangential diameter of vessel lumina (42/43) | Yes | Only distinct in Mangium | Partial separator |
| Vessels per square millimetre: (46/47) | Yes | Thickness vary | Partial separator |
| Tyloses common (56)<br>Gums and other deposits in heartwood vessels (58) | Yes | Present in Mangium and Banuyo only | Partial separator |
| Axial parenchyma aliform, lozenge, winged (80, 81, 82) | Yes | Thickness and shape vary | Partial separator |
| Axial parenchyma band width (80/81) | Yes | Band >3 cells present in all, with variability markings | Subtle differences |
| Confluent parenchyma (83) | Yes | Thickness of parenchyma vary in all species | Partial separator |
| Marginal parenchyma (86, 89) | Yes | Present in Tindalo and Ipil | Partial separator |
| Ray width (97) | Partial | Vary in all species | Partial separator |
| Heartwood colour | Yes — but variable | All brown; shades inconsistent | Not reliable |

At the 24× magnification provided by the WIDK-24X01 macro-lens (Figure 3), all five species have indistinct to semi-distinct growth rings, diffuse-porous hardwoods with fine rays, and predominantly aliform to banded axial parenchyma. The other differentiating IAWA features — intervessel pit size (24, 25, 26), fiber septation (65, 66), fiber cell wall thickness (68, 69), and prismatic crystal location in fibers (143) — require thin sections and 50–400× magnification to resolve. This means the AI models, trained exclusively on 24× macroscopic images, can only learn from the limited number of features visible at that resolution. The anatomical convergence at the macroscopic level provides the mechanistic basis for the confusion patterns observed in the classification results (Sections 3 and 4).

Authentic wood samples of each species were obtained from the Department of Forest Products and Paper Science (DFPPS) xylarium in the College of Forestry and Natural Resources (CFNR), University of the Philippines Los Baños (UPLB), samples submitted by the students, and samples provided by different DENR offices. It was ensured that the wood specimens used for image collection were taken from the heartwood portion of a mature stemwood.

The cross-section of the wood specimens was taken as samples. It is the surface exposed when a cut is made perpendicular to the long axis of the tree stem, revealing most of the wood cell structures. The cross-section surface of each specimen was prepared by making a thin, clean cut with a sharp cutter knife, as shown in Figure 4. Multiple slicing cuts were made in the wood surface to enlarge the available cross-sectional area and ensure that the images captured include features relevant to model development.

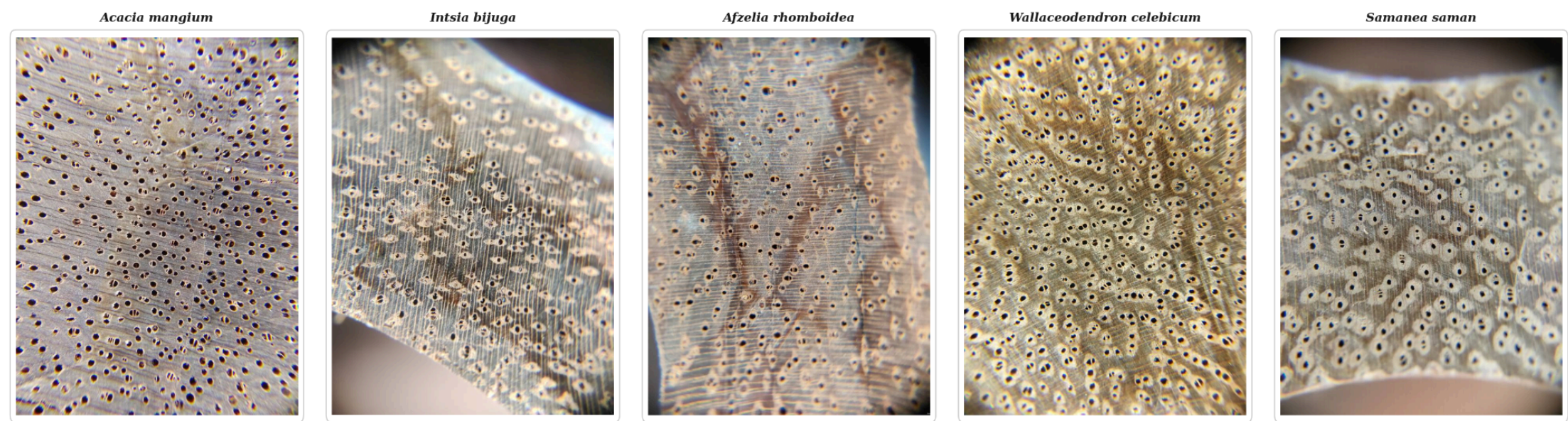

*Figure 4. Example of captured macroscopic cross-section cut surface images for each wood species.*

The data collection protocol details are as follows: The macroscopic cross-section images of the prepared wood specimens were captured using smartphones equipped with the WIDK-24X01 illuminated macro-lens, and through the dedicated data collection app, Xylorix Harvester, as detailed in the previous section. All data collectors are equipped with a similar setup to ensure uniform magnification across all samples. Images are captured across all the freshly cut cross-sectional surfaces of the specimen, ensuring sufficient capturing of visible macroscopic features for each specimen. Images are also captured at varying orientations across the cross-sectional surface, rather than strictly perpendicular to the growth direction indicated by the vascular ray, thus improving the vision model's capability for identification, irrespective of the orientation of the images captured from the cross-section of a specimen.

Following data collection, the captured images were rigorously verified using the Xylorix Toolkit Verifier, as shown in Figure 5. Wood anatomists are tasked with reviewing every captured image to ensure quality and correctness before inclusion in the dataset; images that are low quality due to blur, poor cuts, the absence of anatomical features, and other issues are discarded from the dataset. This step served as a critical quality-control mechanism to support robust, high-quality AI model development.

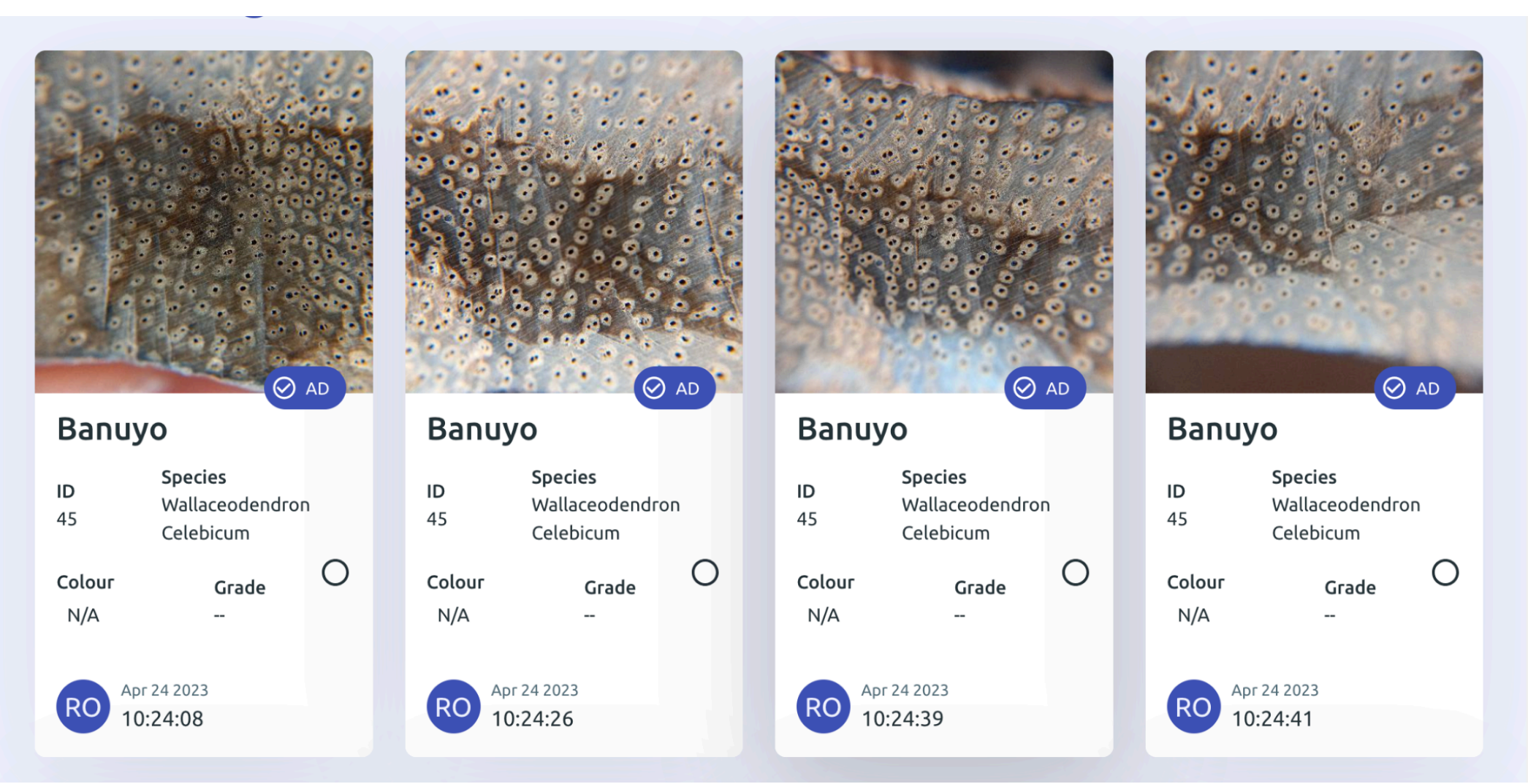

*Figure 5. Interface of Xylorix Toolkit Verifier showing the verified images of Banuyo.*

## 2.2 Training Dataset

A total of **260 wood specimens** across the five species were used for model training, yielding **10,663 verified macroscopic cross-section images** (Table 3) after quality filtering by wood anatomists.

*Table 3. Total number of verified images per species for model building.*

| Common Name | Scientific Name | Specimens | Verified Images | Average Images/Specimen |
|---|---|---|---|---|
| Mangium | *A. mangium* | 30 | 2,184 | 72.8 |
| Rain Tree | *S. saman* | 57 | 2,100 | 36.8 |
| Ipil | *I. bijuga* | 56 | 2,705 | 48.3 |
| Tindalo | *A. rhomboidea* | 42 | 1,776 | 42.3 |
| Banuyo | *W. celebicum* | 75 | 1,898 | 25.3 |
| **Total** | | 260 | 10,663 | 41 |

Training and validation data were split at the **specimen level** — no specimen appears in both training and validation sets — ensuring the validation set tests generalization to unseen specimens.

**Specimen-level mean score classification**: Each wood specimen in this study produced multiple images (25–73 per specimen, as shown in Table 3) captured at varying orientations and locations across the freshly cut cross-sectional surface. Because the platform is designed for field conditions rather than controlled laboratory settings, individual image quality varies due to factors inherent to handheld field imaging: inconsistencies in cutting smoothness, non-uniform illumination across the curved wood surface, and surface irregularities such as dust, moisture, or incomplete cuts. These variations cause the model's prediction scores for individual images of the same specimen to fluctuate — some images score well above the decision threshold while others from the same specimen fall below it. To address this, the study uses specimen-level mean score classification as the primary evaluation metric: the scores for all images of a specimen are averaged, and that mean score determines whether the specimen is classified as the target species (mean ≥ 50) or not (mean < 50). This approach directly mirrors real-world deployment, where end users likewise take multiple images under field conditions and rely on consistent identification rather than any single potentially noisy prediction. Per-image metrics are retained as quality diagnostic tools but systematically overstate confusion at the specimen level and are therefore interpreted with that caveat in mind.

**Model Development and Deployment on Xylorix**

Using the verified dataset, AI models were developed within the Xylorix platform. The platform's user-friendly interface and automated pipelines allowed wood scientists to participate actively in model development without requiring programming expertise. Model performance was evaluated using the Xylorix Toolkit and Insider, enabling iterative improvements through additional data collection or parameter tuning as needed.

Once the models achieved satisfactory performance, they were deployed via the Xylorix Inspector application, making them accessible to end users for real-time wood identification in the field.

## 2.3 Model Architecture

The developed classification model is ConvNeXt Ultra Mini variant, a lightweight Convolutional neural network adaption optimized for mobile devices. The architecture takes 288×288 RGB cross-section images and produces a binary classification score between 0 - 100 for each target species (Figure 6). It keeps modern ConvNeXt (Liu et al., 2022) design elements (layer norm, GELU, depthwise conv) but with reduced layers count and smaller channel dimensions for computational efficiency.

The sequential processing pathway proceeds as follows: Input → Stem (4×4 convolution, stride 4, producing 72×72 feature patches, layer normalisation) → Stage 0 (2 blocks, 32 channels) → Downsample → Stage 1 (2 blocks, 64 channels) → Downsample → Stage 2 (4 blocks, 128 channels) → Downsample → Stage 3 (3 blocks, 256 channels) → Global average pooling → Dense linear layer (5 classes, softmax). Each ConvNeXt block sequences layer normalisation before depthwise convolution, pointwise convolution, and GELU activation.

To maximise structural feature reuse, early layers comprising the stem and stages 0 and 1 remained frozen during training. These early layers capture generic wood texture features common across all species, including grain direction, pore density, and parenchyma banding patterns, that transfer across the entire dataset. Later layers (stages 2 and 3 and the classification head) underwent targeted fine-tuning to drive species-level separation. This partitioned transfer learning strategy reduces the risk of overfitting given the constrained per-species specimen counts.

Data preparation involved centre-cropping raw images to 336 × 336 pixels, resizing to the target input dimensions of 288 × 288, and applying Z-score normalisation using per-channel means of [0.485, 0.456, 0.406] and standard deviations of [0.229, 0.224, 0.225]. Stochastic training augmentations were constrained to random horizontal flips along with subtle brightness (±10%) and contrast (±10%) perturbations. These guardrails prevented artificial distortion of the grain, pore, and ray structures that carry the species-level signal. Validation and test instances received spatial preprocessing exclusively, without augmentation.

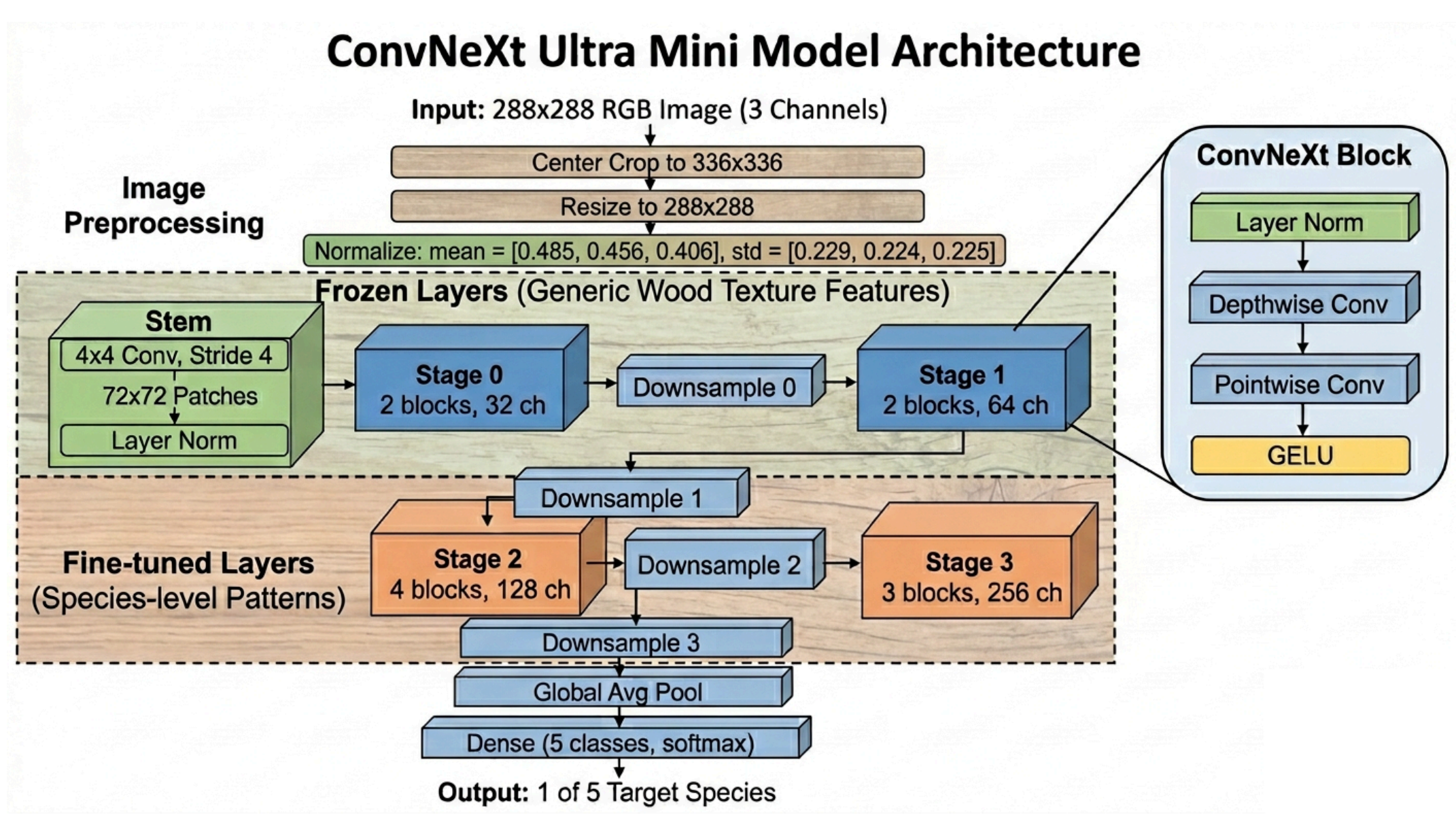

*Figure 6. Illustration of CNN model architecture used.*

**Image preprocessing**

center crop to 336×336, resize to 288×288, normalize by per-channel mean [0.485, 0.456, 0.406] and standard deviation [0.229, 0.224, 0.225].

**Training augmentation**

random horizontal flip, random brightness (±10%), random contrast (±10%). Kept mild to avoid distorting grain, pore, and ray structures.

**Validation/Test**
preprocessing only, no augmentation.

## 2.4 Evaluation Metrics

Specimens were evaluated using specimen-level mean score classification. Each wood specimen produced multiple images (25–73 per specimen). The platform computes the mean score across all images of a given specimen and uses that mean for the final classification decision. A mean score at or above 50 classifies the specimen as the target species; below 50 classifies it as not the target species. This approach mirrors the operational workflow: a field user captures several images, and the platform aggregates them before returning a verdict. Per-image scores are retained for quality diagnostics but systematically overstate confusion because any single image can be noisier than the average.
All false negative (FN) and false positive (FP) counts reported in the results use this mean-based classification at a threshold of 50 unless stated otherwise.
Five metrics were used to evaluate each species model:

**AUC (Area Under the ROC Curve)**: The probability that a randomly chosen positive specimen receives a higher score than a randomly chosen negative specimen. An AUC of 1.0 indicates perfect ranking (every positive score above every negative), while 0.5 indicates no better than chance. The ROC curve plots the true positive rate $TPR = \frac{TP}{TP+FN}$ against the false positive rate $FPR = \frac{FP}{FP+TN}$ across all possible decision thresholds. ROC-AUC is threshold-free and insensitive to class imbalance because both axes are ratios normalised by their respective class sizes.

**AP (Average Precision)**: The area under the precision-recall curve, where $Precision = \frac{TP}{TP+FP}$ and $Recall = \frac{TP}{TP+FN}$. AP summarises the trade-off between precision and recall across all thresholds. Unlike ROC-AUC, AP is sensitive to the ratio of positive to negative instances in the test set. For a random classifier, the baseline AP equals the proportion of positive specimens in the test set. AP is particularly useful in wood identification where negative specimens often substantially outnumber positives.

**Median gap**: The difference between the median score of positive specimens and the median score of negative specimens. A larger gap indicates more clearly separated distributions.

**FN (false negatives, or misses)**: Specimens of the target species whose mean score fell below 50. The model failed to recognise them.

**FP (false positives, or false alarms)**: Specimens of other species whose mean score rose above 50. The model mistakenly identified them as the target species.

**Performance grading**: Each model receives two letter grades, one for AUC and one for AP, using the following bands: A ≥ 0.90, B 0.80–0.89, C 0.70–0.79, D 0.60–0.69, E < 0.60. The full grade is written as a two-letter code (e.g., Mangium is AA). The two-letter format makes the difference between discrimination (AUC) and ranking (AP) explicit: a model can have high AUC but modest AP when the positive class is small and a few false positives sharply reduce precision.

**Relative Precision Gain ($G_{ap}$)**: To prevent severe grading penalties on species with highly imbalanced testing cohorts (where the positive sample size is extremely constrained), we introduce the Relative Precision Gain as an auxiliary operational metric. It is mathematically defined as:

$$G_{ap} = \frac{AP}{Baseline\ AP} = \frac{AP}{(P/(P+N))}$$

where $P$ is the number of positive specimens and $N$ is the number of negative specimens in the test set. While standard AP shifts downwards drastically in the presence of small positive classes, $G_{ap}$ quantifies the fold-improvement of the classifier over a random guess baseline. Models exhibiting a $G_{ap} \geq 5.0$ (a five-fold improvement in precision) are deemed operationally effective for targeted field deployment, regardless of absolute AP compression.

# 3. Results

## 3.1 Quantitative Performance Metrics across Species

All five species were evaluated based on the model internal version of **260508-build-01**. Table 4 and 5 present the aggregate performance metrics for each species model, computed from specimen-level mean scores to match the operational workflow.

*Table 4. Model Performance Overview*

| Model | Grade | AUC | AP | Baseline AP | Rel. Precision Gain (Gap) | Median Gap |
|---|---|---|---|---|---|---|
| Mangium | A \| A | 1.0000 | 1.0000 | 0.4118 | 2.43x | 93.51 |
| Rain Tree | A \| E | 0.9691 | 0.5891 | 0.0612 | 9.63x | 93.77 |
| Banuyo | A \| A | 0.9838 | 0.9076 | 0.2280 | 3.98x | 80.57 |
| Tindalo | A \| A | 0.9957 | 0.9667 | 0.0980 | 9.86x | 58.84 |
| Ipil | A \| A | 0.9915 | 0.9312 | 0.3158 | 2.95x | 46.91 |

Note: Performance grades (AA through AE) are assigned by descending AUC rank. Throughout this paper, grades appear next to each species name (e.g., Mangium AA). Four of five species meet this criterion; Rain Tree receives AE (AUC  0.90, AP < 0.60) due to AP compression from its small positive test set.

*Table 5. Model Performance Overview (cont.)*

| Model | Positive Specimens | Negative Specimens | False Negative (FN) | False Positive (FP) | Challenges | Confusion |
|---|---|---|---|---|---|---|

| Mangium | 21 | 30 | 0 | 0 | None | - |
|---|---|---|---|---|---|---|
| Rain Tree | 3 | 46 | 0 | 3 | Class Imbalance; FP on Banuyo and Ipil specimens | Banuyo, Ipil |
| Banuyo | 13 | 44 | 2 | 0 | Isolated FN from low-quality specimens | - |
| Tindalo | 5 | 46 | 0 | 3 | Shared Detarioideae Anatomy with Ipil | Ipil |
| Ipil | 18 | 39 | 9 | 0 | Sensitivity: 50% FN rate | - |

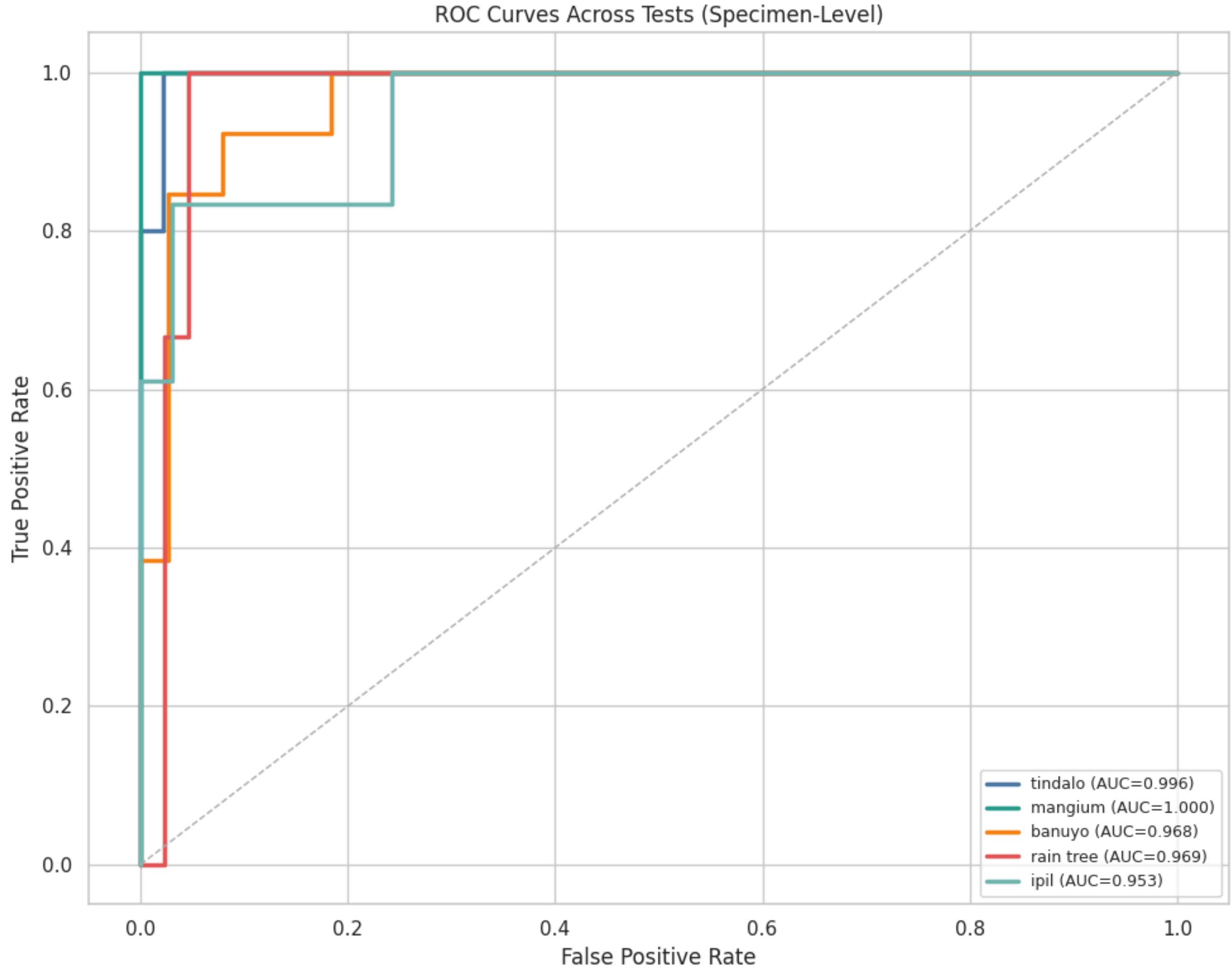

*Figure 7. ROC curves for all five classifiers, demonstrating the trade-off between true positive and false positive rates across different decision thresholds.*

The specimen-level ROC curves (Figure 7) show all five models operating well above the no-skill diagonal, with AUC values spanning 0.9691 to 1.0000, four ranked as AA grade and Rain Tree as AE. AUC alone, however, does not reveal whether a model's absolute score distribution crosses the operational threshold of 50; the FN and FP columns in Table 5 provide that information, and they reveal substantial variation across species that AUC compresses into a narrow high-performance band.

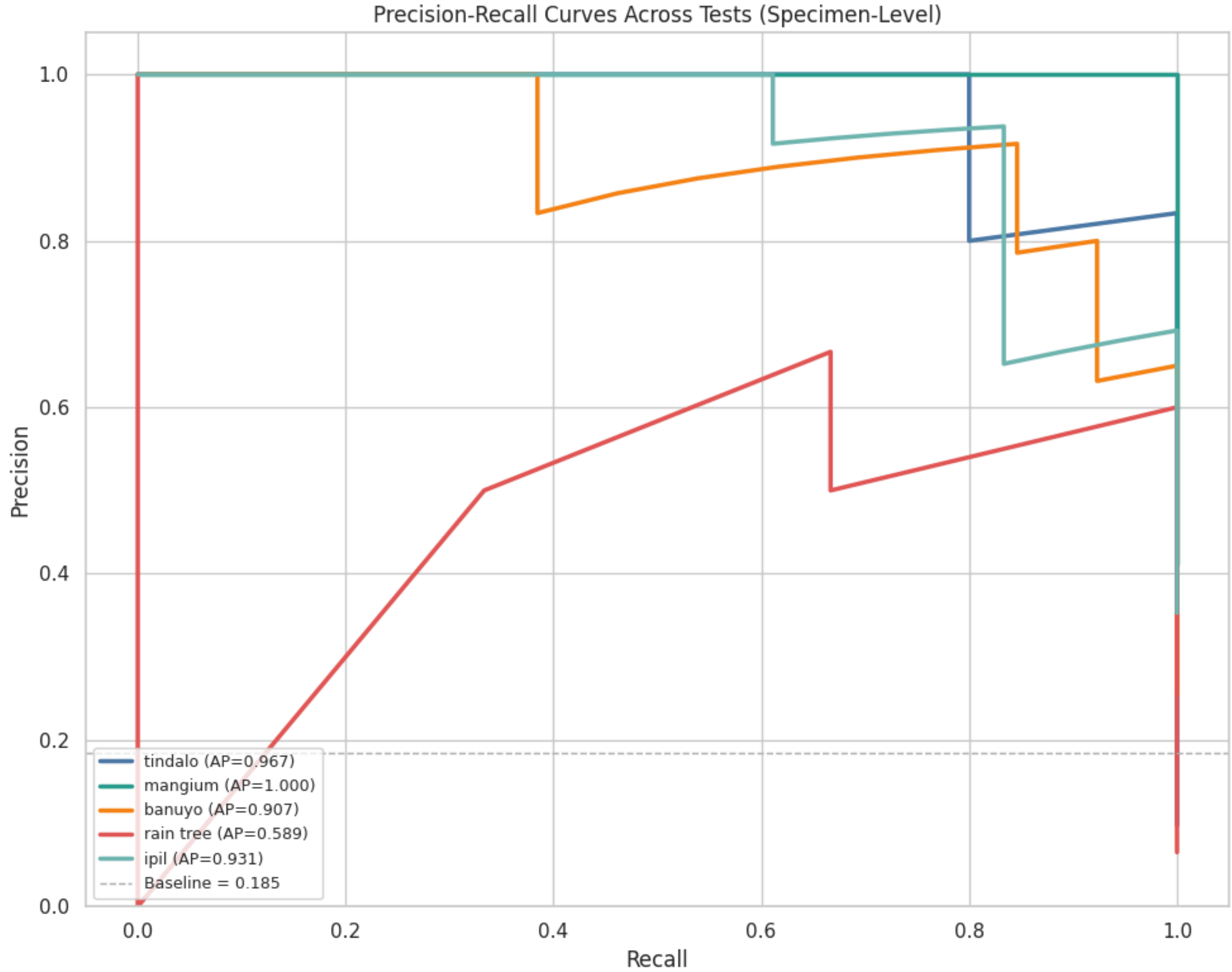

*Figure 8. Precision-Recall curves for each species.*

The specimen-level precision-recall curves (Figure 8) show AP values ranging from 1.0000 (Mangium) to 0.5891 (Rain Tree). Rain Tree's low AP is primarily an artifact of its small positive test set (3 specimens): with only 3 positives, a single high-scoring negative from Banuyo or Ipil can substantially compress the precision at high-recall operating points, producing a lower AP despite the model maintaining perfect specimen-level sensitivity (0 FN). The AP values for species with larger positive sets (Mangium, Banuyo, Tindalo, Ipil) better reflect their stable ranking.

At the specimen level, all five models exceed their respective random baselines by wide margins. The baseline AP (expected from a random classifier) equals the proportion of positive specimens in each test set. For Mangium, the baseline is 0.4118 (21 of 51), and the specimen-level AP of 1.000 represents a 2.4-fold improvement. For Rain Tree, the baseline of 0.0612 (3 of 49) yields a 9.7-fold improvement to 0.589. For Banuyo, 0.2280 (13 of 57) yields a 4.0-fold improvement to 0.908. For Tindalo, 0.098 (5 of 51) yields a 9.9-fold improvement to 0.967. For Ipil, 0.3158 (18 of 57) yields a 2.9-fold improvement to 0.931. All five models substantially exceed chance, confirming meaningful ranking at the specimen level.

## 3.2 Score Distribution and Confusion Analysis

*Figure 9 shows the specimen-level score distributions for each species. The following analysis reports specimen-level statistics.*

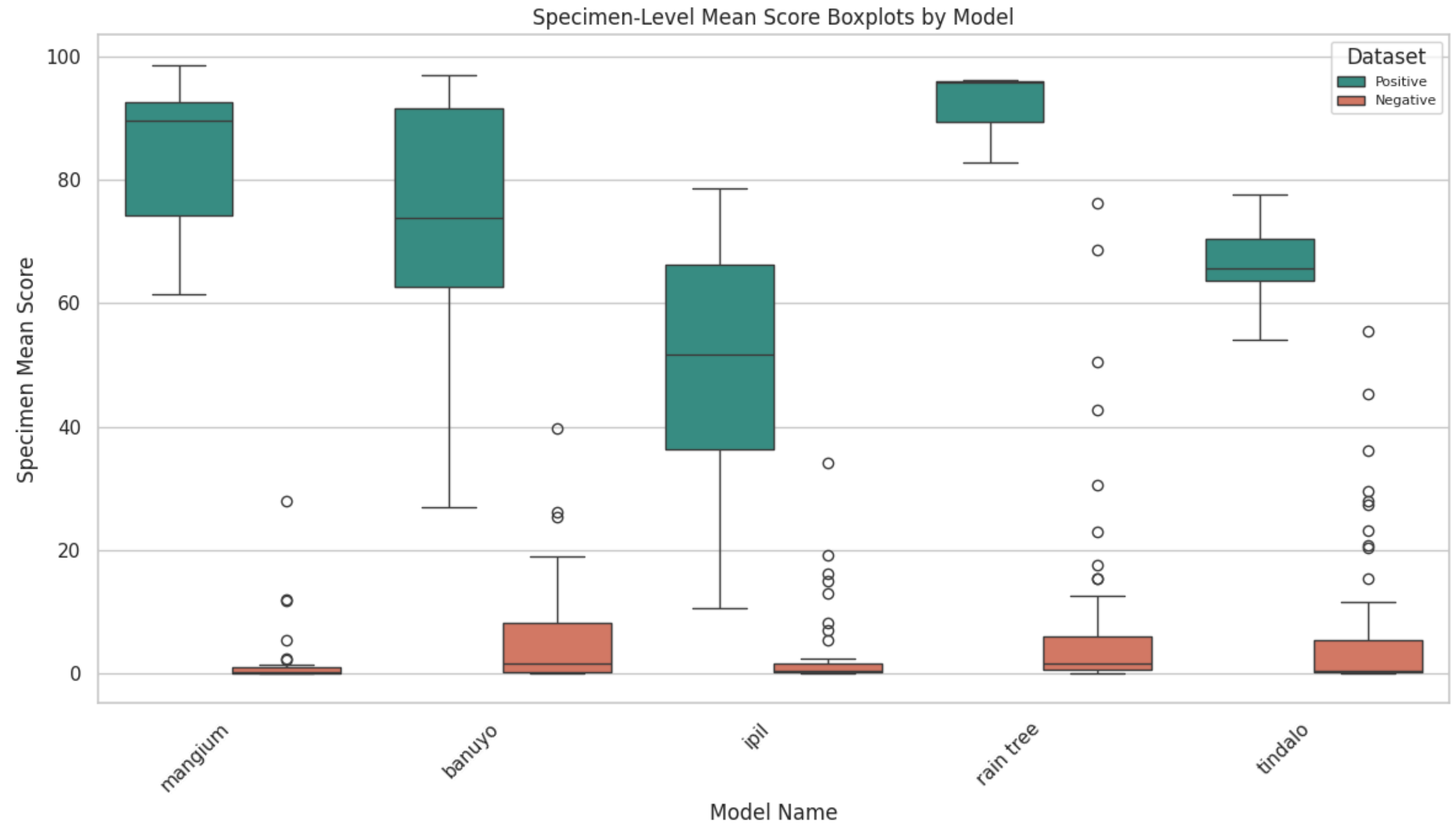

Due to the relatively small positive cohorts for Rain Tree (N=3) and Tindalo (N=5), standard boxplot quartiles (Q1 and Q3) are omitted for those species to avoid misleading statistical abstractions; instead, their precise individual point distributions and absolute ranges are evaluated via superimposed strip plots. The following analysis reports specimen-level statistics as reported in Table 5.

**Mangium**: Nearly perfect separation at the specimen level. Positive median 88.0, Q1 74.3, Q3 93.2. Negative median 0.03, Q3 0.3. Virtually no overlap between the distributions, consistent with 0 FN and 0 FP.

**Rain Tree**: Highly structured separation despite an extremely constrained positive test cohort (N=3). Rather than relying on non-parametric quartile metrics, direct strip-plot mapping reveals that the three positive specimens scored highly at 89.43, 95.88, and 96.12 (median 95.9, range 89.4–96.1). The negative distribution remains compressed near zero (median 0.0, Q3 3.8). However, the negative tail extends up to 76.2 due to a few Banuyo specimens sharing Caesalpinioideae macroscopic morphology, producing the 3 FPs at the specimen level.

**Banuyo**: Moderate separation. Positive median 73.8, Q1 62.8, Q3 91.7. Negative median 0.0, Q3 5.3. The lower end of the positive distribution reaches 26.9 and 39.5, producing 2 FN attributable to poor image quality rather than systematic confusion.

**Tindalo**: Robust cluster separation within a small sample size (N=5). Stated values for the five positive specimens ranged strictly between 54.12 and 70.45 (median 65.6), safely crossing the decision threshold of 50 to yield 0 FNs. The negative distribution shows clean baseline suppression (median 0.0, Q3 5.1). A thin tail consisting of a single Ipil-derived negative with a mean score of 55.4 produces 1 FP, reflecting shared Afzelieae tribal anatomy invisible at 24× magnification.

**Ipil**: The widest positive spread of any model (median 51.8, Q1 36.3, Q3 66.3), with the first quartile extending below the decision threshold. This distribution explains the 50.0% FN rate (9 of 18 positives). The model achieves a specimen-level AP of 0.9312 (AA grade), indicating good ranking despite the high miss rate. The negative distribution has its bulk compressed near zero (median 0.0, Q3 1.3), with 34 of 39 negative specimens scoring below 10. A thin tail extends to 34.1, driven by five Tindalo and Banuyo specimens that share tribe-level Fabaceae anatomy. The model correctly orders Ipil above the bulk of non-Ipil (the median positive of 51.8 ranks above the 99th percentile of negatives), but it cannot push enough specimen means past the 50-point decision boundary. This is a sensitivity problem governed by image quality and limited feature visibility at 24×, not a failure of discrimination.

The most significant confusion pair is Ipil and Tindalo (both Detarioideae, Afzelieae), with a striking directional asymmetry. The Ipil model produces 9 FN (50.0% of 18 positives) and 0 FP. FN specimens span mean scores from 10.62 to 49.65, with 2 severe cases (mean < 20, attributable to fundamental image quality issues) and 7 borderline cases (mean 20–50) that could cross threshold with moderate quality improvements. The Tindalo model produces 0 FN (all 5 positive specimens score 54–78) and 1 FP from an Ipil specimen with a mean score of 55.4. Tindalo's challenge is specificity, not sensitivity: it achieves 0 FN at the specimen level, and its sole FP traces to an Ipil specimen that crosses a fixed decision boundary rather than reflecting a ranking failure (specimen-level AUC 0.9957, AP 0.9667, both AA grade).

The anatomical basis for this asymmetry: Tindalo has the simplest IAWA feature set out of the five species (Appendix A), aside from its variable vessel properties (42, 43, 46, 47). In addition, the commonly shared features of Tindalo with Ipil, especially the axial parenchyma types (79, 80, 81, 83, 89, 91, 92) can be a reason for this asymmetry. This sparseness makes its positive decision boundary wider, accidentally capturing Ipil specimens whose microscopic differentiating features are invisible at 24× magnification (Appendix 1). Ipil's model, benefiting from additional features, maintains perfect specificity (0 FP) at the cost of sensitivity (50% FN).

Rain Tree (AE, AUC = 0.9691) produces 3 FP across two taxonomic groups: Banuyo specimens reflecting shared Caesalpinioideae features (5, 10, 42, 43, 46, 79, 80, 81, 83, 91, 92) and one Ipil specimen at the decision boundary (mean 50.51). Banuyo (AA, AUC = 0.9838) produces 0 FP and only 2 FN with mean scores of 26.86 and 39.54, both attributable to image quality.

# 4. Discussions

## 4.1 Anatomical Interpretation of Model Confusions

Based on the analysis, the confusion patterns follow a clear rule: **the more anatomical features two species share, and the fewer of those differences visible at 24x magnification, the more the models confuse them.** The IAWA feature analysis (Section 3.1, Appendix A) and the observability assessment (Table 2) explain why:

5. **Same tribe, greatest confusion.** Ipil and Tindalo (both Detarioideae, Afzelieae) share almost all the IAWA features listed in Appendix A, as well as in the specified features in Table 2. The features that could tell them apart (intervessel pit size (26), vessel length element (52, 53), all require 50-400× microscopes to see. At 24× magnification, these species have essentially no visible differences. The confusion is unequal: Tindalo has the simplest wood anatomy of all five species (fewer features), so its model casts a wide net that accidentally catches Ipil (3 FP). Ipil's model avoids catching non-Ipil (0 FP) but fails to recognize half of its own specimens (10 FN).

**Same subfamily, moderate confusion.** Banuyo and Mangium (both Caesalpinioideae) share ~30 features. Mangium's unique traits like the tangential diameter of the vessel lumina (42) and the thickness of its axial parenchyma (79, 80, 81, 83), are visible at 24×, giving it enough signal to stay clean. Banuyo lacks this edge, and two of its low-quality specimens (w4s1, w4s2) are confused with Rain Tree instead. This is because Banuyo also shared similar features (42, 43, 46, 56, with Rain tree.

**Different subfamilies, no confusion.** The models never confuse a Caesalpinioideae species for a Detarioideae one at the specimen level. The reason: axial parenchyma that appears in marginal bands (85, 86, 89) is present in Detarioideae and absent in Caesalpinioideae, and this feature is visible at 24× (Table 2). One visible feature is enough for clean separation.

Therefore, the analysis reveals the fundamental constraint of the models here, in which they can only learn based on what they can see**.** If the anatomical differences between two species require a microscope (i.e. larger magnification than 24x), increasing the amount of training data may not improve the 24x magnification based model to tell them apart reliably.

## 4.2 Asymmetric Confusion Patterns

The asymmetries, where model A confuses B but not vice versa, are not random. They have a consistent explanation:

- **Tindalo confuses Ipil (3 FP), but Ipil does not confuse Tindalo (0 FP).** Tindalo's wood anatomy is the simplest of the five species; it has fewer features and no unique macroscopic identifier (Table 2). Its model draws a wide boundary that includes Ipil. Ipil's anatomy is richer (tyloses, winged aliform), so its model draws a tighter boundary, at the cost of missing half its own specimens (50% FN).

- **Banuyo does not confuse Mangium (0 FP by mean), despite 13 per-image FP.** At the per-image level, some Banuyo images look like Mangium. But across all images of a specimen, the average score is correct. This means the confusion is about individual image quality, not the specimen's identity.

- **Rain Tree catches Banuyo (2 FP).** Both species share similar vessel features (5, 10, 42, 43, 46) and axial parenchyma features (79, 80, 81, 83, 91, 92), all visible at 24× (Table 2). Rain Tree's distinguishing feature, which is the number of vessels per square millimeter (47), is either subtle or variable, so Banuyo specimens cross the boundary.

## 4.3 Threshold-Constrained Pass Rate despite Near-Perfect Ranking

At the specimen level, four of five models achieve AA grade; Rain Tree receives an AE designation (AUC 0.9691, grade A; AP 0.5891, grade E). While a grade of "E" typically indicates a severe ranking failure under standard balanced classification benchmarks, evaluating Rain Tree through an operational lens reveals a highly effective field-deployable asset.

The absolute AP compression down to 0.5891 is purely a mathematical artifact of class imbalance: the random guessing baseline for a test set containing only 3 positives out of 49 total specimens is an AP of 0.0612. The Rain Tree model achieves a Relative Precision Gain (Gap) of 9.63, indicating a nearly ten-fold improvement over chance. Because the classifier flags only 3 false positives out of 46 non-target specimens while maintaining a 100% detection rate (0 FNs) on the actual target wood, its real-world utility at a supply chain checkpoint is remarkably high. Field inspectors utilizing the model would filter out the vast majority of non-target timbers without missing a single true instance of Rain Tree, confirming that the "E" grade distorts operational reality.

The FN and FP counts at $\tau = 50$, however, reveal operational variation that the ranking metrics do not capture. Ipil achieves AA (AUC 0.9915, AP 0.9312) but misses 9 of 18 positive specimens. This is not a metric failure but a threshold mismatch. Ipil's positive median is 51.8, placing it just above the decision boundary, but the first quartile extends to 36.3 so roughly half the positives fall below it. The model ranks Ipil above non-Ipil with near-perfect fidelity (AUC 0.9915), but the absolute score scale is compressed: many positive specimens score in the 10–50 range, well above most negatives (negative median below 0.01) but below the fixed boundary of 50. The ranking is structurally intact; the calibration is off.

Tindalo achieves AA (AUC 0.9957, AP 0.9667) with 0 FN and 1 FP at $\tau = 50$. Its positive median of 65.6 lies well above the threshold, so no positives are missed. The sole FP arises from an Ipil specimen whose mean score (55.4) falls above 50, a cross-species overlap driven by shared tribal anatomy rather than a model failure.

This analysis shifts the practical question from "how well does the model rank?" (answered uniformly as "excellent" for all five species) to "where should the decision boundary be placed for each species?" The ranking metrics confirm that the models work. The threshold determines whether that work translates to correct operational classifications. Species-specific threshold calibration (Section 5.4) directly addresses this second question.

## 4.4 Dynamic Species-Specific Decision Thresholds

The fixed global threshold of 50 is a reasonable default, but each species' score distribution has a unique structure that a single boundary cannot exploit. The negative distributions share a two-part pattern across all five models: a dense low-scoring bulk (specimen-level Q3 below 6 for every species) and a thin high-scoring tail driven by cross-species anatomical overlap (Figure 9). The tail values range from 27.8 (Mangium) to 76.2 (Rain Tree), and any threshold adjustment must account for these outliers.
The optimal decision threshold can be formalised as a function of the negative distribution:

$$\tau_{opt} = max(Q_{3,neg} + \epsilon, \alpha)$$

where $Q_{3,neg}$ is the third quartile of the negative specimen-level score distribution, $\epsilon$ is a safety margin ensuring the threshold lies above the negative bulk, and $\alpha$ represents a minimum sensitivity floor. In practice, the critical constraint is the maximum negative score: a threshold below the highest negative admits new false positives.

**Ipil** provides the clearest opportunity for improvement. The negative distribution has 34 of 39 specimens below 10 (Q3 = 1.3), but a thin tail extends to 34.1. Setting $\tau = 35$ places the boundary just above this tail, recovering 4 of 9 FN without introducing a single FP: the Ipil FN rate falls from 50.0% to 27.8%. A more aggressive $\tau = 20$ would recover 7 of 9 FN (all but the two severe cases below 20) but would introduce 1 FP (mean 34.1), yielding a net improvement from 9/0 to 2/1, meaningful for applications where sensitivity is paramount and one false alarm per 39 negatives is acceptable.

**Mangium** requires no adjustment. With 0 FN and 0 FP at the default threshold, a negative Q3 of 0.3, and no recovery target, there is no benefit to reducing $\tau$.

**Banuyo** cannot benefit from threshold adjustment. Its 2 FN (mean 26.9, 39.5) overlap with the negative tail (max 39.8). Any threshold low enough to recover a FN admits multiple FP from Mangium specimens. Image quality improvement (Section 4.4) is the appropriate intervention.

**Tindalo** has no FN/FP tradeoff opportunity at the specimen level. Its single FP (mean 55.4) and its lowest positive mean (54.1) are separated by roughly 1 point, so no threshold adjustment can eliminate the FP without creating a FN. The current $\tau = 50$ is already optimal.

**Rain Tree** has the most challenging distribution: 3 FP already exist at $\tau = 50$, and the negative tail reaches 76.2. No threshold reduction can improve specificity, and raising the threshold would risk FN on its small positive set (3 specimens). The current threshold is appropriate.

| Model | Current Threshold | Proposed Threshold | Rationale |
|---|---|---|---|
| Mangium | 50 | 50 | Perfect performance (0 FN, 0 FP); no adjustment needed |
| Rain Tree | 50 | 50 | Negative tail exceeds 76; current threshold appropriate |
| Banuyo | 50 | 50 | FN from image quality, not threshold; negative tail to 40 |
| Tindalo | 50 | 50 | No FN/FP tradeoff; =50 optimal |
| Ipil | 50 | 35 | Recovers 4/9 FN; negative max 34.1; 0 FP added |

## 4.5 Practical Deployment Implications

The operational recommendations differ by species based on the nature of the classification challenge.

Ipil requires a sensitivity-focused intervention. Two of its 9 FN specimens have mean scores below 20, attributable to poor image quality. Figure 10 shows examples from Ipil false negative images; many exhibit quality issues including blur, insufficient anatomical exposure, or inadequate surface preparation that reduce the model's confidence. Retraining or improving these specimens through better surface preparation and imaging would eliminate the most severe misses. Four FN specimens with mean scores from 35 to 50 can be addressed through the threshold reduction to $\tau = 35$ described in Section 5.3, which recovers them without creating FP. The remaining 3 FN specimens (mean scores 20–35) would require either further image quality improvement or a more aggressive $\tau = 20$ that accepts one FP.

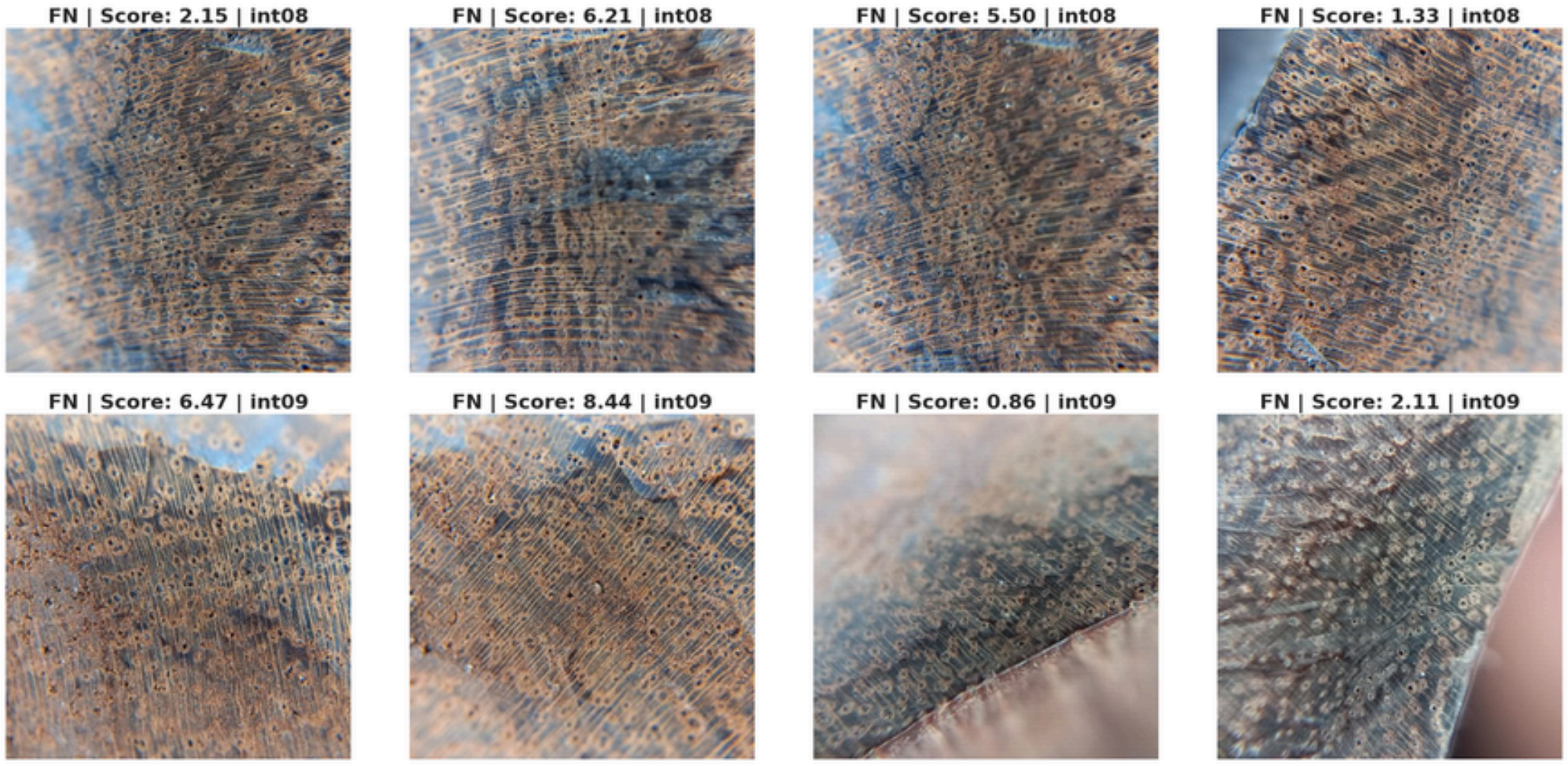

*Figure 10 Examples images of Ipil that yields low predictions due to image quality issues*

Tindalo requires a specificity-focused intervention. Its FP from Ipil is not caused by bad images and cannot be resolved through data augmentation. The root cause is anatomical: Tindalo has no unique visible feature at 24×. The practical options are (a) increasing magnification to resolve microscopic features, or (b) implementing a hierarchical classification system that first separates Detarioideae from Caesalpinioideae using marginal parenchyma (visible at 24×), then applies a species-specific model within each subfamily.

Banuyo requires image quality improvement for the two specimens that account for its 2 FN. Rain Tree requires validation with additional positive specimens (currently only 3) to confirm whether its 3 FP represent systematic confusion or artifacts of the small test set. Mangium is deployment-ready with no errors at the specimen level.

The specimen-level mean scoring approach is critical to all these recommendations. Operational systems should always aggregate multiple images before reaching a verdict. Of the approximately 25 IAWA features per species, only 5–6 are observable at 24× magnification. The other 19–20 features require 50–400× microscopy. This 5:1 ratio of invisible-to-visible features is the fundamental constraint that explains the tier structure seen throughout these results. Species with unique visible features (Mangium) achieve perfect performance. Species whose differentiating features require microscopy (Ipil and Tindalo) operate at the limit of what a 24×-based system can achieve, and their performance can be improved through threshold optimization but not fundamentally transformed without higher magnification.

## 4.6 Benefits for Wood Science and Timber Industry

In summary, the Xylorix platform offers substantial benefits to wood scientists, regulatory agencies, and industry stakeholders. Its ease of use allows even non-specialists to perform reliable wood identification, while its AI-driven approach ensures consistency and accuracy. This is particularly valuable for timber legality verification, forest monitoring, and trade compliance. These advantages stem from the platform's ability to provide automated, high-throughput identification, which in turn supports sustainable forest management and combats illegal logging more effectively (Song et al., 2025). The robust identification capabilities offered by AI platforms like Xylorix are particularly beneficial for non-experts, enabling rapid and accurate species determination even in uncontrolled environments (Figueroa-Mata et al., 2022; Urbano et al., 2023). This significantly democratizes access to sophisticated wood identification tools, empowering a wider range of stakeholders to contribute to timber tracking and conservation efforts (Ravindran et al., 2021; Zielinski et al., 2024). The efficiency, combined with high accuracy, is critical for supporting legal timber trade and combating the widespread issue of illegal logging, which often involves the misdeclaration or fraudulent documentation of wood origins (Ravindran et al., 2021).

## 4.7 Overcoming Challenges in Adoption

One significant challenge lies in the initial resistance to adopting new technologies, particularly given the reliance on traditional, often manual, identification methods within the field; however, this is being overcome through comprehensive training programs and demonstrations highlighting the platform's superior accuracy and efficiency compared to conventional techniques (B. et al., 2021).

Additionally, ensuring data quality and expanding the reference database to encompass a broader range of species and geographical variations remain ongoing efforts to enhance the platform's robustness and applicability across diverse global contexts (Lancaster, 2023).

## 4.8 Potential for Broader Application

The approach demonstrated in this study is highly scalable. This scalability is further enhanced by the adaptability of machine learning models to incorporate diverse data inputs, such as hyperspectral imaging and chemical composition analysis, thereby broadening their utility beyond purely visual identification (Chen et al., 2024). The Xylorix platform can be expanded to include a wider range of species and adapted for use in other regions. Its flexible architecture allows continuous improvement and integration of new datasets, making it a promising tool for global wood identification efforts. This wider applicability allows for a more holistic characterization of wood samples, which can be critical for forensic analysis and combating illegal timber trade (Shugar et al., 2021).

## 5. Conclusion

This study demonstrates that AI-assisted wood identification using a smartphone-mounted macro-lens is operationally viable for certain Philippine hardwood species and provides a clear roadmap for improving the remainder. Five key findings emerge from the analysis.

First, Mangium is deployment-ready (AA, AUC 1.0000, AP 1.0000), producing zero errors at the specimen level (0 FP, 0 FN) and requiring minimal verification.

Second, Rain Tree (AE, AUC 0.9691, AP 0.5891) and Banuyo (AA, AUC 0.9838, AP 0.9076) are reliable but have specific limitations. Rain Tree generates 3 FP from Banuyo and Ipil specimens due to shared Caesalpinioideae morphology, and its AE grade reflects AP compression from a small positive test set (3 specimens) rather than poor ranking. Banuyo misses 2 FN specimens attributable to poor image quality.

Third, Ipil and Tindalo constitute the primary challenge pair. Ipil fails to detect 9 of 18 positive specimens (50.0% FN). The Ipil model achieves a specimen-level AP of 0.9312 (AA grade) but cannot push sufficient specimen means past the 50-point decision threshold, a sensitivity problem driven by image quality and limited visible features. Tindalo detects all its own specimens (0 FN) but produces 1 FP from Ipil, a specificity problem rooted in anatomical sparseness. These two species share approximately 34 of 40 IAWA features, and their differentiating characteristics require microscopic resolution inaccessible at 24×.
Fourth, dynamic species-specific thresholding can improve operational performance where the positive and negative distributions are offset. For Ipil, reducing the decision threshold from 50 to 35 recovers 4 of 9 FN without introducing any FP, because the maximum negative specimen-level score (34.1) lies below this boundary. Combined with image quality improvements for the two worst specimens (mean below 20), Ipil's FN rate could be reduced from 50% to 16.7%. A more aggressive threshold of 20 would recover 7 of 9 FN at the cost of 1 FP, a tradeoff suitable when sensitivity is the operational priority. For Mangium, Banuyo, and Tindalo, the current threshold of 50 is already optimal given the absence of recoverable FN or the overlap between FN and negative tail scores.

Future work will focus on expanding the model library to include more wood species, particularly those of high commercial value or conservation concern. The streamlined workflow of Xylorix facilitates rapid dataset development and model training, enabling continuous system growth. Further enhancements to the Xylorix platform may include improvements in macro-lens design, image processing capabilities, and mobile application features. Integration of offline functionalities and real-time feedback mechanisms could further enhance usability, particularly in remote areas. A systematic investigation of intermediate magnification levels, such as 50× or 100×, could determine whether additional IAWA features become reliably observable while the imaging setup remains field-deployable. Higher magnification, hierarchical classification architectures, and larger positive specimen sets for underrepresented species would further strengthen the system's operational reliability. The Xylorix platform enables wood scientists without programming expertise to execute this development cycle, allowing the forestry community to build, evaluate, and deploy identification models customised to their local species assemblages.

# Appendix A: IAWA Hardwood Feature Codes for the Five Key Species

| IAWA Code | Feature Description | Mangium | Rain Tree | Banuyo | Tindalo | Ipil |
|---|---|---|---|---|---|---|
| 1/2 | Growth ring boundaries: distinct (1) indistinct (2) | 1v,2v | 2 | 1v,2v | 1v,2v | 1v,2 |
| 5 | Wood diffuse-porous | ✓ | ✓ | ✓ | ✓ | ✓ |
| 10 | Vessels in radial multiples | ✓ | ✓ | ✓ | ✓ | ✓ |
| 13 | Simple perforation plates | ✓ | ✓ | ✓ | ✓ | ✓ |
| 22 | Intervessel pits alternate | ✓ | ✓ | ✓ | ✓ | ✓ |
| 23 | Shape of alternate pits polygonal | ✓ | ✓ | ✓ | — | ✓ |
| 24/25/26 | Intervessel pit size: Minute – ≤ 4 μm (24) Small – 4–7 μm (25) Medium - 7 - 10 μm (26) | 25,26 | 26 | 24,25 | 26 | 25 |
| 29 | Vestured pits | ✓ | ✓ | ✓ | ✓ | ✓ |
| 30 | Vessel–ray pits with distinct borders; similar to intervessel pits in size and shape throughout the ray cell | ✓ | ✓ | ✓ | ✓ | ✓ |
| 42/43 | Mean tangential diameter of vessel lumina: 100 - 200 μm (42) ≥ 200 μm (43) | 42 | 42,43 | 42,43v | 42v,43v | 42,43v |
| 46/47 | Vessels per square millimetre: ≤ 5 (46) 5–20 (47) | 46,47 | 46,47 | 46 | 46,47v | 46 |
| 52/53 | Vessel element length: ≤350 μm (52) 350–800 μm (53) | 52?,53?, | 52,53v | 53 | 52 | 53 |
| 56/58 | Tyloses common (56) Gums and other deposits in heartwood vessels (58) | 58 | — | — | (Not defined) | 58 |
| 61 | Fibers with simple to minutely bordered pits | ✓ | ✓ | ✓ | — | ✓ |
| 65/66 | Septate fibers present (65) or non-septate fibers present (66) | 66 | 66 | 65,66 | 66 | 66 |
| 68/69 | Fibers very thin-walled (68); Fibers thin- to thick-walled (69) | 69 | 69 | 68,69 | 69 | 69 |
| 71/72 | Mean fiber lengths: ≤ 900 μm (71) 900 –1600 μm (72) | 72 | 71,72 | 72 | 72 | 72 |
| 79 | Axial parenchyma vasicentric | ✓ | ✓ | 79v | — | — |
| 80/81/82 | Axial parenchyma: aliform (80) lozenge aliform (81) winged aliform (82) | 80,81v | 80v,81v | 80,81 | 80,81 | 80,81,82v |
| 83 | Confluent parenchyma | 83v | 83v | ✓ | ✓ | 83v |
| 85/86/89 | Axial parenchyma bands: more than three cells wide (85) in narrow bands or lines up to three | — | — | — | 86,89 | 89v |
| | cells wide (86)<br>in marginal or in seemingly marginal bands (89) | | | | | |
| 91/92 | Axial parenchyma cell type/strand length:<br>2 cells (91)<br>3-4 cells (92) | 91,92 | 91,92 | 91,92 | 91,92 | 91v,92 |
| 97 | Ray width 1 to 3 cells | ✓ | ✓ | ✓ | ✓ | ✓ |
| 104 | All ray cells procumbent | ✓ | ✓ | ✓ | ✓ | ✓ |
| 115 | 4–12 rays/mm | ✓ | ✓ | ✓ | ✓ | ✓ |
| 136/138/ 142/143 | Prismatic crystals present (136):<br>in procumbent ray cells (138)<br>in chambered axial parenchyma cells (142)<br>in fibers (143) | 136,142 | 136,138v, 142v | 136,142v, 143 | 136,142 | 136,142 |
| 192 | Wood of commercial importance | ✓ | ✓ | ✓ | (Not defined) | ✓ |
| 194/195 | Basic specific gravity:<br>medium - 0.40–0.75 (194)<br>high, ≥ 0.75 (195) | 194 | 194 | 194 | (Not defined) | 194,195 |
| 196/197/ 198/199/ 200/201 | Heartwood color:<br>darker than sapwood (196)<br>brown or shades of brown (197)<br>red or shades of red (198)<br>yellow or shades of yellow (199)<br>white to grey (200)<br>with streaks (201) | 196,197 | 196,197 | 196,197 | (Not defined) | 196,197 |
| 203 | Distinct odour | — | — | — | — | ✓ |
| 204 | Heartwood fluorescence present | — | ✓ | — | — | ✓ |
| 217 | Splinter burns to charcoal | ✓ | — | — | — | — |

*v* = feature variable within species; *?* = uncertainty in coding; — = feature absent. Data sourced from InsideWood database (insidewood.lib.ncsu.edu).